\documentclass{article}

\usepackage{times}
\usepackage{graphicx} 

\usepackage{natbib}

\usepackage{algorithm}
\usepackage{algorithmic}

\usepackage{hyperref}

\RequirePackage{xcolor}
\usepackage{xcolor}
\usepackage[accepted]{icml2018}
\usepackage{times}
\usepackage{caption}
\usepackage{subcaption}
\usepackage{latexsym}
\usepackage{tikz}
\usepackage{comment}
\usetikzlibrary{arrows}
\usetikzlibrary{positioning,shadows.blur}
\usepackage{pifont}

\usepackage{amsmath,amssymb}
\usepackage{xifthen}
\usepackage{graphicx}
\usepackage{graphics}
\usepackage{float}
\usepackage{algorithm}
\usepackage{hyperref}
\usepackage{booktabs}

\newcommand\MLP[0]{\mathrm{MLP}}
\newcommand\GRU[1]{\mathrm{GRU}_{#1}}

\newcommand{\todo}[1]{}
\renewcommand{\todo}[2]{{\color{red} TODO[#1]: {#2}}}
\newcommand{\miketodo}[1]{}
\renewcommand{\miketodo}[1]{{\color{red} Mike: {#1}}}
\newcommand{\denistodo}[1]{}
\renewcommand{\denistodo}[1]{{\color{green} Denis: {#1}}}

\DeclareMathOperator*{\argmax}{\arg\!\max}
\usetikzlibrary{shapes.geometric, arrows, patterns, decorations.pathreplacing}

\newcommand{\rnn}[0]{\textsc{RNN}\ }
\newcommand{\hierarchical}[0]{\textsc{Hierarchical}\ }
\newcommand{\latentintents}[0]{\textsc{Baseline Clusters}\ }
\newcommand{\fullmodel}[0]{\textsc{Full}\ }
\newcommand{\human}[0]{\textsc{Human}\ }

\renewcommand{\vec}[1]{#1}

\icmltitlerunning{Hierarchical Text Generation and Planning for Strategic Dialogue}

\begin{document} 

\twocolumn[
\icmltitle{Hierarchical Text Generation and Planning for Strategic Dialogue}



\icmlsetsymbol{equal}{*}

\begin{icmlauthorlist}
\icmlauthor{Denis Yarats}{equal,fb}
\icmlauthor{Mike Lewis}{equal,fb}

\end{icmlauthorlist}

\icmlaffiliation{fb}{Facebook AI Research, Menlo Park, CA}

\icmlcorrespondingauthor{Denis Yarats}{denisy@fb.com}

\vskip 0.3in
]




\printAffiliationsAndNotice{\icmlEqualContribution}


\begin{abstract}


End-to-end models for goal-orientated dialogue are challenging to train, because linguistic and strategic aspects are entangled in latent state vectors. 
We introduce an approach to learning representations of messages in dialogues by maximizing the likelihood of subsequent sentences and actions, which decouples the semantics of the dialogue utterance from its linguistic realization.
We then use these latent sentence representations for hierarchical language generation, planning and reinforcement learning. Experiments show that our approach  increases the end-task reward achieved by the model, improves the effectiveness of long-term planning using rollouts, and allows self-play reinforcement learning to improve decision making without diverging from human language.
Our hierarchical latent-variable model outperforms previous work both linguistically and strategically.


\end{abstract}

\section{Introduction}

Word-by-word approaches to text generation have been successful in many tasks. However, they have limitations in under-constrained generation settings, such as dialogue response or summarization, where models have significant freedom in the semantics of the text to generate. In such cases, models are prone to overly generic responses that may be valid but suboptimal \citep{li:2015, li:2016, das:2017}.
Further, such models are uninterpretable and somewhat intellectually dissatisfying because they do not cleanly  distinguish between the semantics of language and its surface realization.
Entangling form and meaning is problematic for reinforcement learning, where backpropagating caused by semantic decisions can adversely affect the linguistic quality of text \citep{lewis:2017}, and for candidate generation for longterm planning, as linguistically diverse text may lack semantic diversity.

We focus on negotiation dialogues, where the text generated by the model has consequences than can be easily measured.
Substitutions of similar words (for example substituting a "one" for a "two") can have a large impact on the end-task reward achieved by a dialogue agent.
We use a hierarchical generation approach for a strategic dialogue agent, where the agent first samples a short-term plan in the form of a latent sentence representation. The agent then conditions on this plan during generation, allowing precise and consistent generation of text to achieve a short-term goal. Doing so, we aim to disentangle the concepts of \emph{"what to say"} and \emph{"how to say it"}.
To do this, we introduce a method for learning discrete latent representations of sentences based on their effect on the continuation of the dialogue.


Recent work has explored hierarchical generation of dialogue responses, where a latent variable $z_t$ is inferred to maximize the likelihood of a message $x_t$, given previous messages $x_{0:t-1} \equiv (x_0, \ldots, x_{t-1})$ \citep{serban:2016b, serban:2016c,wen:2017,cao:2017}, which has the effect of clustering similar message strings. Our approach differs in that the latent variable $z_t$ is optimized to maximize the likelihood of messages and actions of the continuation of the dialogue, but not the message $x_t$ itself. Hence, $z_t$ learns to represent $x_t$'s effect on the dialogue, but not the words of $x_t$.The distinction is important because messages with similar words can have very different semantics; and conversely the same meaning can be conveyed with different sentences. 
We show empirically and through human evaluation that our method leads both to better perplexities and end task rewards, and qualitatively that our representations group sentences that are more semantically coherent but linguistically diverse.




We use our message representations to improve the strategic decision making of our dialogue agent. We improve the model's ability to plan ahead by creating a set of semantically diverse candidate messages by sampling distinct $z_t$, and then use rollouts to identify the an expected reward for each. 
We also apply reinforcement learning based on the end-task reward. Previous work has found that RL can adversely effect the fluency of the language generated by the model 
We instead show that simply fine-tuning the parameters for choosing $z_t$ allows the model to substantially improve its rewards while maintaining human-like language.



Experiments show that our approach to disentangling the form and meaning of sentences leads to agents that use language more fluently and intelligently to achieve their goals.

\section{Background}
\subsection{Natural Language Negotiations}
We focus on the negotiation task introduced by \citet{lewis:2017}, as it possess both linguistic and reasoning challenges. Lewis et al. collected a corpus of human dialogues on a multi-issue bargaining task, where the agents must divide a collection of items of 3 different types (\emph{books}, \emph{hats} and \emph{balls}) between them. Actions correspond to choosing a particular subset of the items, and agents choose compatible actions if each item is assigned to exactly one agent.

More formally, the agents $X$ and $Y$ are initially given a space $\mathcal{A}$ of possible agreements, and value functions $v^X$ and $v^Y$, which specify a non-negative reward for each agreement $a\in \mathcal{A}$. Agents cannot directly observe each other's value functions and can only infer it through a dialogue. The agents sequentially exchange turns of natural language $x_t$, consisting of $n_t + 1$ words $x_t^{0:n_t} \equiv (x_t^0, \ldots, x_t^{n_t})$, until one agent enters a special turn that ends the dialogue. Then, both agents independently enter agreements $a^X, a^Y \in \mathcal{A}$ respectively. If the agreements are compatible, both agents receive a reward based on their action and the value function. If the actions are incompatible, neither agent receives any reward.
Training dialogues from an agent's perspective consist of agreement space $\mathcal{A}$, value function $v$, messages $x_{0:T}$ and agreement $a$.

\subsection{Challenges in Text Generation}
We identify a number of challenges for end-to-end text generation for strategic dialogue. These problems have been identified in other text generation settings, but strategic dialogue makes an interesting test case, where decisions have measurable consequences.
\begin{itemize}
    \item \textbf{Lack of semantic diversity}: Multiple samples from a model are often paraphrases of the same intent. This lack of a diversity is a problem if samples are later re-ranked by a long-term planning model.
    \item \textbf{Lack of linguistic diversity}: Neural language models often capture the head of the distribution, providing less varied language than people \citep{li:2015}.
    \item \textbf{Lack of internal coherence}: Messages generated by the model often lack self consistency---for example, \emph{I'll take one hat, and give you all the hats}.
    \item \textbf{Lack of contextual coherence}: Utterances may also lack coherence given the dialogue context so far. For example, \citet{lewis:2017} identify cases where a model starts a message by indicating agreement, but then proposes a counter offer. 
    \item \textbf{Entanglement of linguistic and strategic parameters}: End-to-end approaches do not cleanly distinguish between what to say and how to say it. This is problematic as reinforcement learning aiming to improve decision making may adversely affect the quality of the generated language.
\end{itemize}

We argue that these limitations partly stem from the word-by-word sampling approach to generation, with no explicit plan in advance of generation for what the meaning of the sentence is to be. In \S\ref{section:experiments}, we show our hierarchical approach to generation helps with these problems.
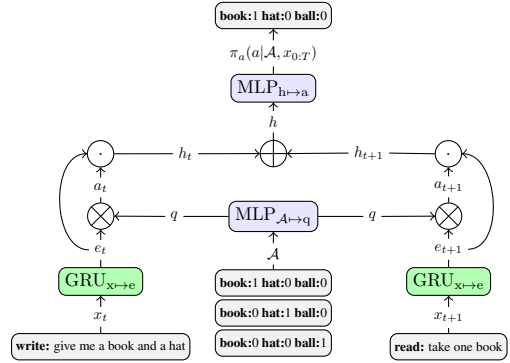
\begin{figure}[t]
\centering
\begin{tikzpicture}[
encoder_node/.style={rectangle, draw=black, fill=green!30, thin, rounded corners=1mm,  minimum width=15mm, minimum height=5mm},
latent_node/.style={diamond, draw=black, fill=blue!40, thin,   text width=1mm},
memory_node/.style={rectangle, draw=black, fill=yellow!20, thin, rounded corners=1mm,  minimum width=15mm, minimum height=5mm},
decoder_node/.style={rectangle, draw=black, fill=yellow!70, thin, rounded corners=1mm,  minimum width=15mm, minimum height=5mm},
selection_node/.style={rectangle, draw=black, fill=red!20, thin, rounded corners=1mm,  minimum width=15mm, minimum height=5mm},
text_node/.style={rectangle, draw=black, fill=gray!10, thin, rounded corners=1mm,  minimum width=15mm, minimum height=7mm},
matrix_node/.style={rectangle, draw=black, fill=blue!10, thin, rounded corners=1mm,  minimum width=5mm, minimum height=5mm},
product_node/.style={circle, draw=black, fill=white, thin, rounded corners=1mm,  minimum width=5mm, minimum height=5mm},
]

\node[encoder_node, scale=0.7] (encoder_x_t) {$\mathrm{GRU_{x \mapsto e}}$};
\node[above=1mm of encoder_x_t.north, scale=0.58]  (e_t) {$e_t$};
\draw[-] (encoder_x_t.north) -- (e_t.south);
\node[below=1.5mm of encoder_x_t.south, scale=0.58]  (x_t) {$x_t$};
\node[text_node, below=5mm of encoder_x_t.south, scale=0.50]  (x_t_text) {\textbf{write:} give me a book and a hat};
\draw[-] (x_t_text.north) -- (x_t.south);
\draw[->] (x_t.north) -- (encoder_x_t.south);
\node[product_node,scale=0.7, above=5mm of encoder_x_t.north] (w_t){};

\draw[-] (w_t.225) -- (w_t.45);
\draw[-] (w_t.135) -- (w_t.315);
\draw[->] (e_t.north) -- (w_t.south);
\node[above=0.7mm of w_t.north, scale=0.58]  (a_t) {$a_t$};

\node[encoder_node, right=35mm of encoder_x_t.east, scale=0.7] (encoder_x_t_1) {$\mathrm{GRU_{x \mapsto e}}$};
\node[above=1mm of encoder_x_t_1.north, scale=0.58]  (e_t_1) {$e_{t+1}$};
\draw[-] (encoder_x_t_1.north) -- (e_t_1.south);
\node[below=1.5mm of encoder_x_t_1.south, scale=0.58]  (x_t_1) {$x_{t+1}$};
\node[text_node, below=5mm of encoder_x_t_1.south, scale=0.50]  (x_t_1_text) {\textbf{read:} take one book};
\draw[-] (x_t_1_text.north) -- (x_t_1.south);
\draw[->] (x_t_1.north) -- (encoder_x_t_1.south);
\node[product_node,scale=0.7, above=5mm of encoder_x_t_1.north] (w_t_1){};
\draw[-] (w_t_1.225) -- (w_t_1.45);
\draw[-] (w_t_1.135) -- (w_t_1.315);
\draw[->] (e_t_1.north) -- (w_t_1.south);
\node[above=0.7mm of w_t_1.north, scale=0.58]  (a_t_1) {$a_{t+1}$};

\node[matrix_node, above right=4.7mm and 17mm of encoder_x_t.north, scale=0.7] (aenc) {$\mathrm{MLP_{\mathcal{A} \mapsto q}}$};
\node[below=1.5mm of aenc.south, scale=0.58]  (a) {$\mathcal{A}$};

\node[text_node, below=5mm of aenc.south, scale=0.50]  (a_1) {\textbf{book:}1 \textbf{hat:}0 \textbf{ball:}0};
\node[text_node, below=9mm of aenc.south, scale=0.50]  (a_2) {\textbf{book:}0 \textbf{hat:}1 \textbf{ball:}0};
\node[text_node, below=13mm of aenc.south, scale=0.50]  (a_3) {\textbf{book:}0 \textbf{hat:}0 \textbf{ball:}1};
\draw[-] (a_1.north) -- (a.south);
\draw[->] (a.north) -- (aenc.south);

\node[left=6mm of aenc.west, scale=0.58]  (a_left) {$q$};
\draw[-] (aenc.west) -- (a_left.east);
\draw[->] (a_left.west) -- (w_t.east);
\node[right=6mm of aenc.east, scale=0.58]  (a_right) {$q$};
\draw[-] (aenc.east) -- (a_right.west);
\draw[->] (a_right.east) -- (w_t_1.west);

\node[product_node,scale=0.7, above=4.5mm of aenc.north] (q){};
\draw[-] (q.90) -- (q.270);
\draw[-] (q.0) -- (q.180);

\node[product_node,scale=0.7, above left=5mm and 21.7mm of aenc.north] (z_t){$\mathbf{.}$};
\path[->] (e_t.west) edge [out=180, in=180] (z_t.west);
\draw[-] (w_t.north) -- (a_t.south);
\draw[->] (a_t.north) -- (z_t.south);

\node[product_node,scale=0.7, above right=5mm and 22mm of aenc.north] (z_t_1){$\mathbf{.}$};
\path[->] (e_t_1.east) edge [out=0, in=0] (z_t_1.east);

\node[above left=4.7mm and 10mm of aenc.north, scale=0.58]  (h_t) {$h_t$};
\draw[-] (z_t.east) -- (h_t.west);
\draw[->] (h_t.east) -- (q.west);

\node[above right=4.7mm and 10mm of aenc.north, scale=0.58]  (h_t_1) {$h_{t+1}$};
\draw[-] (w_t_1.north) -- (a_t_1.south);
\draw[->] (a_t_1.north) -- (z_t_1.south);
\draw[->] (h_t_1.west) -- (q.east);
\draw[-] (z_t_1.west) -- (h_t_1.east);

\node[above=1mm of q.north, scale=0.58]  (h) {$h$};
\node[matrix_node, above=5mm of q.north, scale=0.7] (clas) {$\mathrm{MLP_{h \mapsto a}}$};
\draw[-] (q.north) -- (h.south);
\draw[->] (h.north) -- (clas.south);

\node[text_node, above=6mm of clas.north, scale=0.50]  (out) {\textbf{book:}1 \textbf{hat:}0 \textbf{ball:}0};
\node[above=1mm of clas.north, scale=0.58]  (prob) {$\pi_{a}(a|\mathcal{A}, x_{0:T})$};
\draw[-] (clas.north) -- (prob.south);
\draw[->] (prob.north) -- (out.south);

\end{tikzpicture}
\caption{\label{figure:classifier}Action classifier $\pi_{a}(a|\mathcal{A}, x_{0:T})$, which predicts distribution over actions using a GRU with attention.}

\end{figure}

\section{Action Classifier}
\label{model:classifier}
Initially, we train an \emph{action classifier} $\pi_a(a|\mathcal{A}, x_{0:T})$ (Figure \ref{figure:classifier}) that predicts the final action chosen at the end of the dialogue. This classifier is used in all versions of our model. We implement the action classifier as an RNN with attention \citep{bahdanau:2014}. We first encode the set of possible actions $\mathcal{A}$ as $q = \MLP_{\mathcal{A} \mapsto s}(\mathcal{A})$, and each sentence $x_t$ as $e_t = \GRU{x \mapsto e}(E x_t^{0:n_t})$. We then acquire a fixed size representation $h$ by applying the following transformations:
$a_t = e_t * q, h_t = e_t \odot a_t, h = \sum_{t=0}^T h_t$.
Finally, we apply a softmax classifier:
\begin{align*}
 \pi_{a}(a|\mathcal{A}, x_{0:T}) &\propto \exp(\MLP_{h \mapsto a}(h)) \nonumber
\end{align*}
We train this network to minimize the negative log likelihood of an action $a$ given a set of possible actions $\mathcal{A}$ and a dialogue $x_{0:T}$:
\begin{align*}
\mathcal{L} &= -\sum_{a, \mathcal{A}, x_{0:T}} \log \pi_{a}(a | \mathcal{A}, x_{0:T}) \nonumber
\end{align*}

\section{Baseline Hierarchical Model}
\label{model:baseline}
\begin{figure}[t]
\centering

\begin{tikzpicture}[
pred_encoder_node/.style={rectangle, draw=black, fill=green!20, thin, rounded corners=1mm,  minimum width=15mm, minimum height=5mm},
latent_node/.style={diamond, draw=black, fill=blue!40, thin,   text width=1mm},
memory_node/.style={rectangle, draw=black, fill=green!70, thin, rounded corners=1mm,  minimum width=15mm, minimum height=5mm},
encoder_node/.style={rectangle, draw=black, fill=yellow!20, thin, rounded corners=1mm,  minimum width=15mm, minimum height=5mm},
decoder_node/.style={rectangle, draw=black, fill=yellow!70, thin, rounded corners=1mm,  minimum width=15mm, minimum height=5mm},
selection_node/.style={rectangle, draw=black, fill=yellow!20, thin, rounded corners=1mm,  minimum width=15mm, minimum height=5mm},
text_node/.style={rectangle, draw=black, fill=gray!10, thin, rounded corners=1mm,  minimum width=15mm, minimum height=7mm},
matrix_node/.style={rectangle, draw=black, fill=blue!10, thin, rounded corners=1mm,  minimum width=5mm, minimum height=5mm},
]

\node[pred_encoder_node, scale=0.7] (encoder_x_t) {$\mathrm{GRU_{x \mapsto e}}$};
\node[below=1.5mm of encoder_x_t.south, scale=0.58]  (x_t) {$x_t$};
\node[text_node, below=5mm of encoder_x_t.south, scale=0.50]  (x_t_text) {\textbf{write:} give me 2 hats};
\node[above=1mm of encoder_x_t.north, scale=0.58]  (e_t) {$e_t$};
\draw[-] (x_t_text.north) -- (x_t.south);
\draw[->] (x_t.north) -- (encoder_x_t.south);
\draw[-] (encoder_x_t.north) -- (e_t.south);

\node[memory_node, above=8.5mm of encoder_x_t.north, scale=0.7] (m_t) {$\mathrm{GRU_{eq \mapsto s}}$};
\node[above=1mm of m_t.north, scale=0.58]  (s_t) {$s_t$};
\draw[->] (e_t.north) -- (m_t.south);
\draw[-] (m_t.north) -- (s_t.south);

\node[decoder_node, above right=8mm and 4mm of m_t.north, scale=0.7] (decoder_x_t_1) {$\mathrm{GRU_{s \mapsto x}}$};
\path[->] (s_t.north) edge [out=90, in=180] (decoder_x_t_1.west);
\node[below=1.5mm of decoder_x_t_1.south, scale=0.58]  (x_t_1_up) {$x_{t+1}$};
\draw[->] (x_t_1_up.north) -- (decoder_x_t_1.south);

\node[above=1mm of decoder_x_t_1.north, scale=0.58]  (p_x_t_1) {$p_x(x_{t+1}|\mathcal{A}, v, x_{0:t})$};
\node[text_node, above=6mm of decoder_x_t_1.north, scale=0.50]  (hat_x_t_1_text) {\textbf{read:} take one};
\draw[-] (decoder_x_t_1.north) -- (p_x_t_1.south);
\draw[->] (p_x_t_1.north) -- (hat_x_t_1_text.south);

\node[matrix_node, right=2.5mm of encoder_x_t.east, scale=0.7] (aenc) {$\mathrm{MLP_{\mathcal{A}v \mapsto q}}$};
\node[below=1.5mm of aenc.south, scale=0.58]  (a) {$\mathcal{A},v$};
\draw[->] (a.north) -- (aenc.south);
\node[above=1mm of aenc.north, scale=0.58]  (q) {$q$};
\draw[-] (aenc.north) -- (q.south);

\node[pred_encoder_node, right=18mm of encoder_x_t.east, scale=0.7] (encoder_x_t_1) {$\mathrm{GRU_{x \mapsto e}}$};
\node[below=1.5mm of encoder_x_t_1.south, scale=0.58]  (x_t_1) {$x_{t+1}$};
\node[text_node, below=5mm of encoder_x_t_1.south, scale=0.5]  (x_t_1_text) {\textbf{read:} take one};
\node[above=1mm of encoder_x_t_1.north, scale=0.58]  (e_t_1) {$e_{t+1}$};
\draw[-] (x_t_1_text.north) -- (x_t_1.south);
\draw[->] (x_t_1.north) -- (encoder_x_t_1.south);
\draw[-] (encoder_x_t_1.north) -- (e_t_1.south);

\node[memory_node, above=8.5mm of encoder_x_t_1.north, scale=0.7] (m_t_1) {$\mathrm{GRU_{eq \mapsto s}}$};
\node[above=1mm of m_t_1.north, scale=0.58]  (s_t_1) {$s_{t+1}$};
\draw[->] (e_t_1.north) -- (m_t_1.south);
\draw[-] (m_t_1.north) -- (s_t_1.south);

\path[->] (q.north) edge [out=90, in=270] (m_t.south);
\path[->] (q.north) edge [out=90, in=270] (m_t_1.south);

\node[decoder_node, above right=8mm and 4mm of m_t_1.north, scale=0.7] (decoder_x_t_2) {$\mathrm{GRU_{s \mapsto x}}$};
\path[->] (s_t_1.north) edge [out=90, in=180] (decoder_x_t_2.west);
\node[below=1.5mm of decoder_x_t_2.south, scale=0.58]  (x_t_2_up) {$x_{t+2}$};
\draw[->] (x_t_2_up.north) -- (decoder_x_t_2.south);

\node[above=1mm of decoder_x_t_2.north, scale=0.58]  (p_x_t_2) {$p_x(x_{t+2}|\mathcal{A},v, x_{0:t+1})$};
\node[text_node, above=6mm of decoder_x_t_2.north, scale=0.50]  (hat_x_t_2_text) {\textbf{write:} ok deal};
\draw[-] (decoder_x_t_2.north) -- (p_x_t_2.south);
\draw[->] (p_x_t_2.north) -- (hat_x_t_2_text.south);

\node[left=2mm of m_t.west, scale=0.558]  (m_t_left) {$s_{t-1}$};
\draw[->] (m_t_left.east) -- (m_t.west);
\node[left=1mm of m_t_left.west, scale=0.58]  (dots_left) {$\ldots$};
\draw[-] (dots_left.east) -- (m_t_left.west);

\node[right=6mm of m_t.east, scale=0.58]  (m_t_middle) {$s_t$};
\draw[-] (m_t.east) -- (m_t_middle.west);
\draw[->] (m_t_middle.east) -- (m_t_1.west);

\node[right=1mm of m_t_1.east, scale=0.58]  (m_t_1_right) {$s_{t+1}$};
\draw[-] (m_t_1.east) -- (m_t_1_right.west);
\node[right=1.5mm of m_t_1_right.east, scale=0.58]  (dots_right) {$\ldots$};
\draw[->] (m_t_1_right.east) -- (dots_right.west);


\end{tikzpicture}
\caption{\label{figure:baseline}Baseline hierarchical model.}

\end{figure}
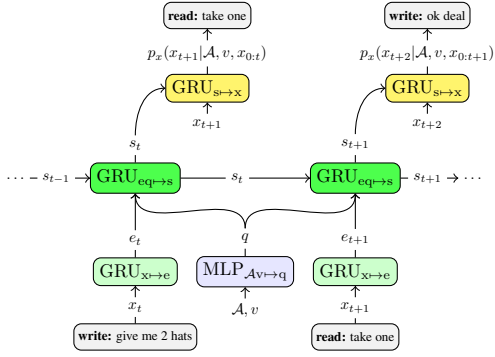

As a baseline, we train a hierarchical encoder-decoder model (Figure \ref{figure:baseline}) to maximize the likelihood of training dialogue sentences, similarly to \citet{serban:2016a}.
The model contains the action-value encoder  to input the action space $\mathcal{A}$ and the value function $v$ as $q = \MLP_{\mathcal{A}v \mapsto q}(\mathcal{A}, v)$; a sentence encoder  that embeds individual messages $x_t$ as $e_t = \GRU{x \mapsto e}(E x_t^{0:n_t})$; a sentence level encoder  that reads sentence embeddings $e_{0:t}$ and the action space encoding $q$ to produce dialogue state $s_t =\GRU{eq  \mapsto s}(e_{0:t}, q)$; and a decoder $\GRU{s \mapsto x}$ that produces message $x_{t+1}$, conditioning on $s_t$: 
\begin{align*}
 p_{x}(x_{t+1}^i | \mathcal{A},v, x_{t+1}^{0:i-1}, x_{0:t}) &\propto \exp (E^\top \GRU{s  \mapsto x}( x_{t+1}^{0:i-1}; s_{t}))  \nonumber\\
 p_{x}(x_{t+1} |\mathcal{A},v, x_{0:t}) &= \prod_{i=0}^{n_{t+1}}p_{x}(x_{t+1}^i |\mathcal{A}, v, x_{t+1}^{0:i-1}, x_{0:t}) \nonumber
\end{align*}
The encoder and decoder share a word embedding matrix $E$.
We minimize the following loss, over the training set:
\begin{align*}
\mathcal{L} &= -\sum_{\mathcal{A},v, x_{0:T}} \sum_{t=0}^T  \log p_{x}(x_t | \mathcal{A}, v,x_{0:t-1})
\end{align*}

\section{Learning Latent Message Representations}
\label{model:cluster}
The central part of our model is a method for encoding messages $x_t$ as discrete latent variables $z_t$. The goal of this model is to learn message representations that reflect the message's effect on the dialogue, but abstract over semantically equivalent paraphrases. The discrete nature of the latent variables $z_t$  allows us to efficiently make sequential decisions by choosing $z_t$  at each step to govern the outcome of the dialogue. We show that such approach is helpful for planning and reinforcement learning.

\begin{figure*}[t!]
\begin{subfigure}[t]{0.5\textwidth}
\centering
\begin{tikzpicture}[
encoder_node/.style={rectangle, draw=black, fill=green!30, thin, rounded corners=1mm,  minimum width=15mm, minimum height=5mm},
latent_node/.style={diamond, draw=black, fill=blue!40, thin,   text width=1mm},
memory_node/.style={rectangle, draw=black, fill=yellow!20, thin, rounded corners=1mm,  minimum width=15mm, minimum height=5mm},
decoder_node/.style={rectangle, draw=black, fill=yellow!70, thin, rounded corners=1mm,  minimum width=15mm, minimum height=5mm},
selection_node/.style={rectangle, draw=black, fill=red!20, thin, rounded corners=1mm,  minimum width=15mm, minimum height=5mm},
text_node/.style={rectangle, draw=black, fill=gray!10, thin, rounded corners=1mm,  minimum width=15mm, minimum height=7mm},
matrix_node/.style={rectangle, draw=black, fill=blue!10, thin, rounded corners=1mm,  minimum width=5mm, minimum height=5mm},
]

\node[encoder_node, scale=0.7] (encoder_x_t) {$\mathrm{GRU_{x \mapsto e}}$};
\node[above=1mm of encoder_x_t.north, scale=0.58]  (e_t) {$e_t$};
\node[matrix_node,scale=0.7, above=7mm of encoder_x_t.north] (w_z_t){$\mathrm{W_{eq \mapsto z}}$};
\node[above=1mm of w_z_t.north, scale=0.58]  (p_z_t_x_t) {$p_z(z_t | \mathcal{A}, x_t)$};
\node[latent_node,scale=0.7, above=17mm of encoder_x_t.north] (latent_z_t){};
\node[above=1mm of latent_z_t.north, scale=0.58]  (z_t) {$z_t$};
\node[below=1.5mm of encoder_x_t.south, scale=0.58]  (x_t) {$x_t$};
\node[text_node, below=5mm of encoder_x_t.south, scale=0.5]  (x_t_text) {\textbf{write:} give me 2 hats};

\node[memory_node, scale=0.7, above=6mm of latent_z_t.north] (memory_h_t) {$\mathrm{GRU_{z \mapsto s}}$};
\node[left=1.5mm of memory_h_t.west, scale=0.58]  (h_t_minus_1) {$s_{t-1}$};
\node[above=2mm of memory_h_t.north, scale=0.58]  (h_t_up) {$s_{t}$};
\node[decoder_node, scale=0.7, above right=11.5mm and 3mm of memory_h_t.north] (decoder_x_t_1) {$\mathrm{GRU_{s \mapsto x}}$};
\node[below=1.5mm of decoder_x_t_1.south, scale=0.58]  (x_t_1_up) {$x_{t+1}$};
\node[above=1mm of decoder_x_t_1.north, scale=0.58]  (p_x_t_1) {$p_x(x_{t+1}| \mathcal{A},z_{0:t})$};
\node[selection_node, scale=0.7, above left=11.5mm and 3mm of memory_h_t.north] (selection_a_t) {$\mathrm{MLP_{s \mapsto a}}$};
\node[above=1mm of selection_a_t.north, scale=0.58]  (p_a_t) {$p_a(a_t| \mathcal{A},z_{0:t})$};
\node[text_node, above=6mm of selection_a_t.north, scale=0.5]  (hat_a_t) {\textbf{book:}0 \textbf{hat:}2 \textbf{ball:}0};
\node[text_node, above=6mm of decoder_x_t_1.north, scale=0.5]  (hat_x_t_1) {\textbf{read:} take one};

\node[encoder_node, scale=0.7, right=25mm of encoder_x_t.east] (encoder_x_t_1) {$\mathrm{GRU_{x \mapsto e}}$};

\node[above=1mm of encoder_x_t_1.north, scale=0.58]  (e_t_1) {$e_{t+1}$};
\node[matrix_node,scale=0.7, above=7mm of encoder_x_t_1.north] (w_z_t_1){$\mathrm{W_{eq \mapsto z}}$};
\node[above=1mm of w_z_t_1.north, scale=0.58]  (p_z_t_1_x_t_1) {$p_z(z_{t+1} | \mathcal{A}, x_{t+1})$};
\node[latent_node, scale=0.7, above=17mm of encoder_x_t_1.north] (latent_z_t_1) {};
\node[above=1mm of latent_z_t_1.north, scale=0.58]  (z_t_1) {$z_{t+1}$};

\node[below=1.5mm of encoder_x_t_1.south, scale=0.58]  (x_t_1) {$x_{t+1}$};
\node[text_node, below=5mm of encoder_x_t_1.south, scale=0.5]  (x_t_1_text) {\textbf{read:} take one};
\node[memory_node,scale=0.7, above=6mm of latent_z_t_1.north] (memory_h_t_1) {$\mathrm{GRU_{z \mapsto s}}$};
\node[right=11mm of memory_h_t.east, scale=0.58]  (h_t_middle) {$s_{t}$};
\node[right=1mm of memory_h_t_1.east, scale=0.58]  (h_t_1_middle) {$s_{t+1}$};
\node[left=1mm of h_t_minus_1.west, scale=0.58]  (dots_left) {$\ldots$};
\node[right=1.5mm of h_t_1_middle.east, scale=0.58]  (dots_right) {$\ldots$};
\node[above=2mm of memory_h_t_1.north, scale=0.58]  (h_t_1_up) {$s_{t+1}$};
\node[decoder_node, scale=0.7, above right=11.5mm and 3mm of memory_h_t_1.north] (decoder_x_t_2) {$\mathrm{GRU_{s \mapsto x}}$};
\node[below=1mm of decoder_x_t_2.south, scale=0.58]  (x_t_2_up) {$x_{t+2}$};
\node[above=1mm of decoder_x_t_2.north, scale=0.58]  (p_x_t_2) {$p_x(x_{t+2}| \mathcal{A},z_{0:t+1})$};
\node[selection_node, scale=0.7, above left=11.5mm and 3mm of memory_h_t_1.north] (selection_a_t_1) {$\mathrm{MLP_{s \mapsto a}}$};
\node[above=1mm of selection_a_t_1.north, scale=0.58]  (p_a_t_1) {$p_a(a_{t+1}| \mathcal{A},z_{0:t+1})$};
\node[text_node, above=6mm of selection_a_t_1.north, scale=0.5]  (hat_a_t_1) {\textbf{book:}0 \textbf{hat:}1 \textbf{ball:}0};
\node[text_node, above=6mm of decoder_x_t_2.north, scale=0.5]  (hat_x_t_2) {\textbf{write:} ok deal};

\node[matrix_node, right=7mm of encoder_x_t.east, scale=0.7] (aenc) {$\mathrm{MLP_{\mathcal{A} \mapsto q}}$};
\node[below=1.5mm of aenc.south, scale=0.58]  (a) {$\mathcal{A}$};
\draw[->] (a.north) -- (aenc.south);
\node[above=1mm of aenc.north, scale=0.58]  (q) {$q$};
\draw[-] (aenc.north) -- (q.south);
\path[->] (q.north) edge [out=90, in=270] (w_z_t.south);
\path[->] (q.north) edge [out=90, in=270] (w_z_t_1.south);

\draw[-] (x_t_text.north) -- (x_t.south);
\draw[->] (x_t.north) -- (encoder_x_t.south);
\draw[-] (encoder_x_t.north) -- (e_t.south);
\draw[->] (e_t.north) -- (w_z_t.south);
\draw[-] (w_z_t.north) -- (p_z_t_x_t.south);
\draw[->] (p_z_t_x_t.north) -- (latent_z_t.south);
\draw[-] (latent_z_t.north) -- (z_t.south);
\draw[->] (z_t.north) -- (memory_h_t.south);
\draw[->] (h_t_minus_1.east) -- (memory_h_t.west);
\draw[-] (memory_h_t.north) -- (h_t_up.south);
\draw[->] (x_t_1_up.north) -- (decoder_x_t_1.south);
\path[->] (h_t_up.north) edge [out=90, in=180] (decoder_x_t_1.west);
\path[->] (h_t_up.north) edge [out=90, in=0] (selection_a_t.east);
\draw[-] (decoder_x_t_1.north) -- (p_x_t_1.south);
\draw[-] (selection_a_t.north) -- (p_a_t.south);
\draw[->] (p_x_t_1.north) -- (hat_x_t_1.south);
\draw[->] (p_a_t.north) -- (hat_a_t.south);

\draw[-] (x_t_1_text.north) -- (x_t_1.south);
\draw[->] (x_t_1.north) -- (encoder_x_t_1.south);
\draw[-] (encoder_x_t_1.north) -- (e_t_1.south);
\draw[->] (e_t_1.north) -- (w_z_t_1.south);
\draw[-] (w_z_t_1.north) -- (p_z_t_1_x_t_1.south);
\draw[->] (p_z_t_1_x_t_1.north) -- (latent_z_t_1.south);
\draw[-] (latent_z_t_1.north) -- (z_t_1.south);
\draw[->] (z_t_1.north) -- (memory_h_t_1.south);
\draw[-] (memory_h_t.east) -- (h_t_middle.west);
\draw[->] (h_t_middle.east) -- (memory_h_t_1.west);
\draw[-] (memory_h_t_1.east) -- (h_t_1_middle.west);
\draw[->] (h_t_1_middle.east) -- (dots_right.west);
\draw[-] (dots_left.east) -- (h_t_minus_1.west);
\draw[-] (memory_h_t_1.north) -- (h_t_1_up.south);
\draw[->] (x_t_2_up.north) -- (decoder_x_t_2.south);
\path[->] (h_t_1_up.north) edge [out=90, in=180] (decoder_x_t_2.west);
\path[->] (h_t_1_up.north) edge [out=90, in=0] (selection_a_t_1.east);
\draw[-] (decoder_x_t_2.north) -- (p_x_t_2.south);
\draw[-] (selection_a_t_1.north) -- (p_a_t_1.south);
\draw[->] (p_x_t_2.north) -- (hat_x_t_2.south);
\draw[->] (p_a_t_1.north) -- (hat_a_t_1.south);

\draw[draw=orange, line width=0.7mm, opacity=0.5] (0, 2.2) circle [x radius=3mm, y radius=5mm];
\draw[draw=orange, line width=0.7mm, opacity=0.5] (3.6, 2.2) circle [x radius=3mm, y radius=5mm];

\end{tikzpicture}
\caption{\label{figure:model:cluster}Clustering Model}
\end{subfigure}
~
\begin{subfigure}[t]{0.5\textwidth}
\begin{tikzpicture}[
pred_encoder_node/.style={rectangle, draw=black, fill=green!20, thin, rounded corners=1mm,  minimum width=15mm, minimum height=5mm},
latent_node/.style={diamond, draw=black, fill=blue!40, thin,   text width=1mm},
memory_node/.style={rectangle, draw=black, fill=green!70, thin, rounded corners=1mm,  minimum width=15mm, minimum height=5mm},
encoder_node/.style={rectangle, draw=black, fill=yellow!20, thin, rounded corners=1mm,  minimum width=15mm, minimum height=5mm},
decoder_node/.style={rectangle, draw=black, fill=yellow!70, thin, rounded corners=1mm,  minimum width=15mm, minimum height=5mm},
selection_node/.style={rectangle, draw=black, fill=yellow!20, thin, rounded corners=1mm,  minimum width=15mm, minimum height=5mm},
text_node/.style={rectangle, draw=black, fill=gray!10, thin, rounded corners=1mm,  minimum width=15mm, minimum height=7mm},
matrix_node/.style={rectangle, draw=black, fill=blue!10, thin, rounded corners=1mm,  minimum width=5mm, minimum height=5mm},
]

\node[pred_encoder_node, scale=0.7] (encoder_x_t) {$\mathrm{GRU_{x \mapsto e}}$};
\node[below=1.5mm of encoder_x_t.south, scale=0.58]  (x_t) {$x_t$};
\node[text_node, below=5mm of encoder_x_t.south, scale=0.50]  (x_t_text) {\textbf{write:} give me 2 hats};
\node[above=1mm of encoder_x_t.north, scale=0.58]  (e_t) {$e_t$};
\draw[-] (x_t_text.north) -- (x_t.south);
\draw[->] (x_t.north) -- (encoder_x_t.south);
\draw[-] (encoder_x_t.north) -- (e_t.south);

\node[memory_node, above=5.5mm of encoder_x_t.north, scale=0.7] (m_t) {$\mathrm{GRU_{eq \mapsto s}}$};
\node[above=1mm of m_t.north, scale=0.58]  (s_t) {$s_t$};
\node[matrix_node,scale=0.7, above=5mm of m_t.north] (w_z_t){$\mathrm{W_{s \mapsto z }}$};
\draw[->] (e_t.north) -- (m_t.south);

\node[latent_node, above=15mm of m_t.north, scale=0.7] (latent_z_t_1){};
\node[above=1mm of w_z_t.north, scale=0.58]  (p_z_t_1_m_t) {$\hat{p}_z(z_{t+1} | \mathcal{A}, v, x_{0:t})$};
\draw[-] (m_t.north) -- (s_t.south);
\draw[->] (s_t.north) -- (w_z_t.south);
\draw[-] (w_z_t.north) -- (p_z_t_1_m_t.south);
\draw[->] (p_z_t_1_m_t.north) -- (latent_z_t_1.south);

\node[above=1mm of latent_z_t_1.north, scale=0.58]  (z_t_1) {$z_{t+1}^*$};
\draw[-] (latent_z_t_1.north) -- (z_t_1.south);

\node[decoder_node, above right=13mm and 4mm of latent_z_t_1.north, scale=0.7] (decoder_x_t_1) {$\mathrm{GRU_{hz \mapsto x}}$};
\path[->] (z_t_1.north) edge [out=90, in=180] (decoder_x_t_1.west);
\node[below=1.5mm of decoder_x_t_1.south, scale=0.58]  (x_t_1_up) {$x_{t+1}$};
\draw[->] (x_t_1_up.north) -- (decoder_x_t_1.south);

\node[encoder_node, below left=6mm and 8mm of decoder_x_t_1.west, scale=0.7] (lmencoder_x_t) {$\mathrm{GRU_{x \mapsto h}}$};
\node[above=1mm of lmencoder_x_t.north, scale=0.58]  (h_t_up) {$h_{t}$};
\draw[-] (lmencoder_x_t.north) -- (h_t_up.south);
\path[->] (h_t_up.north) edge [out=90, in=180] (decoder_x_t_1.west);
\node[below=1.5mm of lmencoder_x_t.south, scale=0.58]  (lm_x_t) {$x_{t}$};
\draw[->] (lm_x_t.north) -- (lmencoder_x_t.south);

\node[above=1mm of decoder_x_t_1.north, scale=0.58]  (p_x_t_1) {$p_x(x_{t+1}|z_{t+1}^*, x_{0:t})$};
\node[text_node, above=6mm of decoder_x_t_1.north, scale=0.50]  (hat_x_t_1_text) {\textbf{read:} take one};
\draw[-] (decoder_x_t_1.north) -- (p_x_t_1.south);
\draw[->] (p_x_t_1.north) -- (hat_x_t_1_text.south);

\node[pred_encoder_node, right=23mm of encoder_x_t.east, scale=0.7] (encoder_x_t_1) {$\mathrm{GRU_{x \mapsto e}}$};
\node[below=1.5mm of encoder_x_t_1.south, scale=0.58]  (x_t_1) {$x_{t+1}$};
\node[text_node, below=5mm of encoder_x_t_1.south, scale=0.5]  (x_t_1_text) {\textbf{read:} take one};
\node[above=1mm of encoder_x_t_1.north, scale=0.58]  (e_t_1) {$e_{t+1}$};
\draw[-] (x_t_1_text.north) -- (x_t_1.south);
\draw[->] (x_t_1.north) -- (encoder_x_t_1.south);
\draw[-] (encoder_x_t_1.north) -- (e_t_1.south);

\node[memory_node, above=5.5mm of encoder_x_t_1.north, scale=0.7] (m_t_1) {$\mathrm{GRU_{eq \mapsto s}}$};
\node[above=1mm of m_t_1.north, scale=0.58]  (s_t_1) {$s_{t+1}$};
\node[matrix_node,scale=0.7, above=5mm of m_t_1.north] (w_z_t_1){$\mathrm{W_{s \mapsto z}}$};
\draw[->] (e_t_1.north) -- (m_t_1.south);

\node[latent_node, above=15mm of m_t_1.north, scale=0.7] (latent_z_t_2){};
\node[above=1mm of w_z_t_1.north, scale=0.58]  (p_z_t_2_m_t_1) {$\hat{p}_z(z_{t+2} |\mathcal{A}, v, x_{0:t+1})$};
\draw[-] (m_t_1.north) -- (s_t_1.south);
\draw[->] (s_t_1.north) -- (w_z_t_1.south);
\draw[-] (w_z_t_1.north) -- (p_z_t_2_m_t_1.south);
\draw[->] (p_z_t_2_m_t_1.north) -- (latent_z_t_2.south);

\node[above=1mm of latent_z_t_2.north, scale=0.58]  (z_t_2) {$z_{t+2}^*$};
\draw[-] (latent_z_t_2.north) -- (z_t_2.south);

\node[decoder_node, above right=13mm and 4mm of latent_z_t_2.north, scale=0.7] (decoder_x_t_2) {$\mathrm{GRU_{hz \mapsto x}}$};
\path[->] (z_t_2.north) edge [out=90, in=180] (decoder_x_t_2.west);
\node[below=1.5mm of decoder_x_t_2.south, scale=0.58]  (x_t_2_up) {$x_{t+2}$};
\draw[->] (x_t_2_up.north) -- (decoder_x_t_2.south);

\node[encoder_node, below left=6mm and 8mm of decoder_x_t_2.west, scale=0.7] (lmencoder_x_t_1) {$\mathrm{GRU_{x \mapsto h}}$};
\node[above=1mm of lmencoder_x_t_1.north, scale=0.58]  (h_t_1_up) {$h_{t+1}$};
\draw[-] (lmencoder_x_t_1.north) -- (h_t_1_up.south);
\path[->] (h_t_1_up.north) edge [out=90, in=180] (decoder_x_t_2.west);
\node[below=1.5mm of lmencoder_x_t_1.south, scale=0.58]  (lm_x_t_1) {$x_{t+1}$};
\draw[->] (lm_x_t_1.north) -- (lmencoder_x_t_1.south);

\node[above=1mm of decoder_x_t_2.north, scale=0.58]  (p_x_t_2) {$p_x(x_{t+2}|z_{t+2}^*, x_{0:t+1})$};
\node[text_node, above=6mm of decoder_x_t_2.north, scale=0.50]  (hat_x_t_2_text) {\textbf{write:} ok deal};
\draw[-] (decoder_x_t_2.north) -- (p_x_t_2.south);
\draw[->] (p_x_t_2.north) -- (hat_x_t_2_text.south);

\node[left=10mm of m_t.west, scale=0.558]  (m_t_left) {$s_{t-1}$};
\draw[->] (m_t_left.east) -- (m_t.west);
\node[left=1mm of m_t_left.west, scale=0.58]  (dots_left) {$\ldots$};
\draw[-] (dots_left.east) -- (m_t_left.west);

\node[right=9mm of m_t.east, scale=0.58]  (m_t_middle) {$s_t$};
\draw[-] (m_t.east) -- (m_t_middle.west);
\draw[->] (m_t_middle.east) -- (m_t_1.west);

\node[right=1mm of m_t_1.east, scale=0.58]  (m_t_1_right) {$s_{t+1}$};
\draw[-] (m_t_1.east) -- (m_t_1_right.west);
\node[right=1.5mm of m_t_1_right.east, scale=0.58]  (dots_right) {$\ldots$};
\draw[->] (m_t_1_right.east) -- (dots_right.west);

\node[left=1.5mm of lmencoder_x_t.west, scale=0.58]  (h_t_minus_1) {$h_{t-1}$};
\draw[->] (h_t_minus_1.east) -- (lmencoder_x_t.west);
\node[left=1mm of h_t_minus_1.west, scale=0.58]  (h_dots_left) {$\ldots$};
\draw[-] (h_dots_left.east) -- (h_t_minus_1.west);

\node[right=10mm of lmencoder_x_t.east, scale=0.58]  (h_t) {$h_{t}$};
\draw[-] (lmencoder_x_t.east) -- (h_t.west);
\draw[->] (h_t.east) -- (lmencoder_x_t_1.west);

\node[right=9mm of lmencoder_x_t_1.east, scale=0.58]  (h_t_1) {$h_{t+1}$};
\draw[-] (lmencoder_x_t_1.east) -- (h_t_1.west);
\node[right=1.5mm of h_t_1.east, scale=0.58]  (h_dots_right) {$\ldots$};
\draw[->] (h_t_1.east) -- (h_dots_right.west);

\node[matrix_node, right=5mm of encoder_x_t.east, scale=0.7] (aenc) {$\mathrm{MLP_{\mathcal{A}v \mapsto q}}$};
\node[below=1.5mm of aenc.south, scale=0.58]  (a) {$\mathcal{A}, v$};
\draw[->] (a.north) -- (aenc.south);
\node[above=1mm of aenc.north, scale=0.58]  (q) {$q$};
\draw[-] (aenc.north) -- (q.south);
\path[->] (q.north) edge [out=90, in=270] (m_t.south);
\path[->] (q.north) edge [out=90, in=270] (m_t_1.south);


\end{tikzpicture}
\caption{\label{figure:model:full}Full Model}
\end{subfigure}
\caption{\label{figure:model}
We pre-train a model to learn a discrete encoder for sentences, which bottlenecks the message $x_t$ through discrete representation $z_t$ (Figure \ref{figure:model:cluster}; \S\ref{model:cluster}). This architecture forces $z_t$ to capture the most relevant aspects of $x_t$ for predicting future messages and actions. We then extract the learned discrete representations $z_t$ (marked by orange ellipses) and train our full model (Figure \ref{figure:model:full}): $p_x$ is trained to translate representations $z^*_t$ into messages $x_t$ (\S\ref{section:lm}), and $\hat{p}_z$ is trained to predict a distribution over $z_t$ given the dialogue history (\S\ref{section:pred}).
 }
\end{figure*}
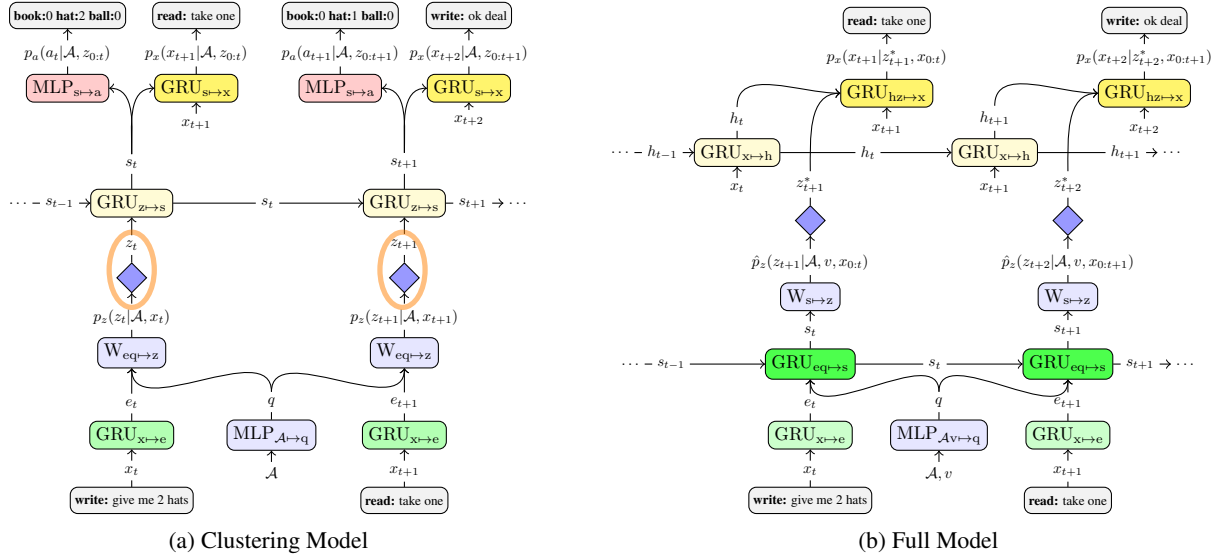

Our representation learning model  (Figure \ref{figure:model:cluster}) has a similar structure to that of \S\ref{model:baseline}, except that message embedding $e_t$ is used as input to a stochastic node $p_z(z_t|\mathcal{A}, x_t)$ formed by a softmax with parameters $W_{eq \mapsto z}$ over latent states $z_t$. We use expectation maximization to learn how to assign messages to clusters to maximize the likelihood of future messages and actions.

After each message $x_t$, $\GRU{z \mapsto s}$ is updated with representation $z_t$ to give hidden state $s_t$. From $s_t$, we train the model to predict the next message $x_{t+1}$ and an action $a_t$. In the training dialogues, there is only an action after the final turn $x_T$; for other turns $x_t$, we use a soft proxy action by regressing to the distribution over actions predicted by $a_t=\pi_a(a|\mathcal{A}, x_{0:t})$. Therefore, $a_t$ is a distribution over what deal would be agreed if the dialogue stopped after message $x_t$. This action can be thought of as latent proxy for a traditional annotated dialogue state \citep{williams:2013}. When predicting $x_{t+1}$ and $a_t$, the model only has access to latent variables $z_{0:t}$, so $z_t$ must contain useful information about the meaning of $x_t$.
We employ a hierarchical RNN, in which message $ e_t = \GRU{x \mapsto e}(Ex_t^{0:n_t})$ and the action space $q = \MLP_{\mathcal{A} \mapsto q}(\mathcal{A})$ encodings are passed through a discrete bottleneck:
\begin{align*}
 p_z(z_t |\mathcal{A}, x_t) &\propto \exp(W_{eq \mapsto z}[e_t, q])  \nonumber\\
 s_t &= \GRU{z \mapsto s}(z_{0:t})  \nonumber\\
 p_x(x_{t+1}^i | \mathcal{A},x_{t+1}^{0:i-1}, z_{0:t}) &\propto \exp (E^\top \GRU{s \mapsto x}(x_{t+1}^{0:i-1};s_t)) \nonumber\\
 p_x(x_{t+1} |\mathcal{A}, z_{0:t}) &= \prod_{i=0}^{n_{t+1}} p_x(x_{t+1}^i |\mathcal{A}, x_{t+1}^{0:i-1}, z_{0:t}) \nonumber\\
 p_a(a_t |\mathcal{A}, z_{0:t}) &\propto \exp(\MLP_{s \mapsto a}(s_t))  \nonumber
\end{align*}
We minimize the following loss over the training set:
\begin{align*}
\mathcal{L} =& \sum_{\mathcal{A}, x_{0:T}} \sum_{t=0}^T  -\log p_{x}(x_{t+1} |\mathcal{A}, z_{0:t}) \\
+& \mathrm{D_{KL}}\left( p_a(a_t |\mathcal{A}, z_{0:t}) || \pi_a(a|\mathcal{A}, x_{0:t}) \right) 
\end{align*}



We optimize latent variables $z$ using minibatch Viterbi Expectation Maximization \cite{dempster:1977}. For each minibatch, for each timestep $t$, we compute:
\begin{align*}
z_t^*=\argmax_z \log( p(x_{t+1}, a_t|\mathcal{A}, z, z_{0:t-1})p_z(z| \mathcal{A},x_t))
\end{align*}
The $\argmax$ requires a separate forward pass for each $z$.
We then advance to the next timestep using $z_t^*$ to update $\GRU{z \mapsto s}$, and finally perform an update maximizing: 
\begin{align*}
\sum_{t=0}^T \log (p(x_{t+1}, a_{t}|\mathcal{A}, z^*_t,z_{0:t-1})p_z(z_t^*|\mathcal{A}, x_{t})) \nonumber
\end{align*}
At convergence, we extract message representations $z_t^*$.






\section{Hierarchical Text Generation}
\label{model:full}

We then train a new hierarchical dialogue model (Figure \ref{figure:model:full}), which uses pre-trained representations $z^{*}_t$ to predict messages $x_t$.
First, we train a recurrent neural network to predict $p_{x}(x_{t+1} | z^{*}_{t+1}, x_{0:t})$. $p_{x}$ learns how to translate the latent variables into fluent text in context. Then, we optimize a model $\hat{p}_z(z_{t+1}|\mathcal{A}, v, x_{0:t})$ to maximize the marginal likelihood of training sentences.



\subsection{Conditional Language Model}
\label{section:lm}
We train $p_{x}$ to translate pretrained representation $z^{*}_t$ and encodings of previous messages $h_t = \GRU{x \mapsto h}(Ex_{0:t})$ into a message $x_t$:
\begin{multline*}
p_{x}(x_{t+1}^i| z_{t+1}^{*},x_{0:t}, x_{t+1}^{0:i-1}) \\\propto \exp(E^{\top}\GRU{hz \mapsto x}(x_{t+1}^{0:i-1};z^{*}_{t+1},h_t))\\
p_x(x_{t+1}|z_{t+1}^{*}, x_{0:t}) = \prod_{i=0}^{n_{t+1}} p_{x}(x_{t+1}^i| z_{t+1}^{*}, x_{0:t}, x_{t+1}^{0:i-1})
\end{multline*}
By minimizing the following loss:
\begin{align*}
\mathcal{L}_x &= -\sum_{x_{0:T}} \sum_{t=0}^T  \log p_x(x_{t+1}| z_{t+1}^{*}, x_{0:t})
\end{align*}



Unlike the baseline model, text generation does not condition explicitly on the agent's value function $v$, or the action space $\mathcal{A}$ -- all knowledge of the goals and available actions is bottlenecked through the dialogue state. This restriction forces the text generation to depend strongly on $z_t$.

\subsection{Latent Variable Prediction Model}
\label{section:pred}
At test time, $z^*_t$ is not available, as it contains information about the future dialogue. Instead, we train a model $\hat{p}_{z}$ to predict $z^*_t$ conditioned on the current dialogue context $s_t = \GRU{e \mapsto s}(e_{0:t}, q)$, where $_t = \GRU{x \mapsto e}(Ex_t^{0:n_t})$ and $q = \MLP_{\mathcal{A}v \mapsto q}(\mathcal{A}, v)$:
\begin{align*}
\hat{p}_z (z_{t+1}|\mathcal{A}, v, x_{0:t}) &\propto \exp (W_{s \mapsto z} s_t)
\end{align*} 
We optimize $\hat{p}_z$ to maximize the marginal likelihood of training messages, without updating $p_{x}$. The model learns to reconstruct the distribution over $z_t$ that best explains message $x_t$.
\begin{align*}
P(x_t|\mathcal{A}, v, x_{0:t-1}) &= \sum_{z} p_{x}(x_t | z, x_{0:t-1})\hat{p}_z(z|\mathcal{A}, v, x_{0:t-1})\\
\mathcal{L}_z &= -\sum_{\mathcal{A}, v, x_{0:T}} \sum_{t=0}^T \log P(x_t|\mathcal{A}, v, x_{0:t-1})
\end{align*} 




\subsection{Decoding}

To generate an utterance $x_t$, the model first samples a predicted plan $z_t$ from $\hat{p}_{z}$:
\begin{align*}
    z_{t} \sim \hat{p}_{z}(z | \mathcal{A}, v, x_{0:t-1})
\end{align*}
The model then sequentially generates tokens $x_t^{i}$  based on plan $z_t$ and context $x_{0:t}$: 
\begin{align*}
    x_t^{i+1} \sim p_{x}(x | z_t, x_{0:t-1}, x_t^{0:i})
\end{align*}
\section{Hierarchial Reinforcement Learning}
\citet{lewis:2017} experiment with end-to-end reinforcement learning to fine-tune pre-trained supervised models. The model engages in a dialogue with another model, achieving reward $V$. This reward is then backpropagated using policy gradients.
One challenge is that because model parameters govern both strategic and linguistic aspects of generation, backpropagating errors can adversely affect the quality of the generated language.
To avoid divergence from human language, we experiment with fixing all model parameters, except for the parameters of $\hat{p}_z$. This allows reinforcement learning to improve decisions about what to say, without affecting language generation parameters. A similar approach was taken in a different dialogue setting by \citet{wen:2017}. 


\section{Hierarchical Planning}
\citet{lewis:2017} propose planning in dialogue using rollouts. First, a set of $K$ unique candidate messages $\{x_t^{(1)}, x_t^{(2)}, \ldots, x_t^{(K)}\}$ are sampled from $p_{x}(x_t|\mathcal{A}, v, x_{0:t-1})$. Then, multiple rollouts of the future dialogue are sampled from the model, and outcomes $a$ are scored according to the value function $v$, to estimate the expected reward $V(x_t)$: 
\begin{align}
V(x_t)=\mathbb{E}_{x_{t+1:T}\sim p_x, a \sim \pi_a}[r(\vec{a}, \vec{v})\pi_a(\vec{a}|\mathcal{A},x_{0:T})]
\label{equation:rollout_reward}
\end{align}
The expectation is approximated with $N$ samples, and the candidate $x^{*}$ with the highest expected score is returned.
\begin{align}
x^{*}=\argmax_{x} V(x)
\label{equation:optimal_message1}
\end{align}
One challenge is that even though the candidates $\{x_t^{(1)}, x_t^{(2)}, \ldots, x_t^{(K)}\}$ can be constrained to be different strings, it is difficult to enforce semantic diversity. For example, if all the candidates are paraphrases of the same intent, then the choice makes little difference to the outcome of the dialogue.
In order to improve the diversity of candidate generation, we take a hierarchical approach of first sampling $K$ unique latent intents $\{z_t^{(1)}, z_t^{(2)}, \ldots, z_t^{(K)}\}$ from $\hat{p}_z(z_t|\mathcal{A}, v, x_{0:t-1})$. Then, for each $z_t^{(i)}$, we choose a candidate turn conditioned on that state: 
\begin{align*}
x_t^{(i)}=\argmax_x p_{x}(x | z_t^{(i)}, x_{0:t-1})
\end{align*}
We then estimate the reward of the candidate message using Equation \ref{equation:rollout_reward}, and finally choose a message as in Equation \ref{equation:optimal_message1}.

\section{Experiments}
\label{section:experiments}

\subsection{Training Details}
We used the following hyper-parameters:: embeddings and hidden states have 256 dimensions; for each unique agreement space $\mathcal{A}$ we learn 50 discrete latent message representations.  During training, we optimize the parameters using RMSProp \citep{tieleman:2012} with initial learning rate 0.0005 and momentum $\mu=0.1$, clipping of gradients whose $L^2$ norm exceeds 1. We train the models for 15 epochs with mini-batch size of 16. We then pick the best snapshot according to validation perplexity and anneal the learning rate by a factor of 5 each epoch. For RL, we use a smaller learning rate of 0.0001, and a discount factor $\gamma$ of 0.95.
For supervised learning we tuned based on validation perplexity; for RL we measured the average reward in self-play.

\subsection{Baselines}
We compare the following models:
\begin{itemize}
    \item \rnn A simple word-by-word approach to generation, similar to \citet{lewis:2017}.
    \item \hierarchical Baseline model in which the two levels of RNN are connected directly, with no discrete bottleneck (\S\ref{model:baseline}), similarly to \citet{serban:2016a}. 
    \item \latentintents Our model (Figure \ref{figure:model:full}) without pretraining the sentence encoder. A latent representation $z_t$ of message $x_t$ is inferred to maximize the likelihood of $p(x_{t+1}, a_t |\mathcal{A}, z_t, z_{0:t-1})p(z_t |\mathcal{A}, x_{t})$. This model is closely related to the \textsc{Latent Intents Dialogue Model} \citep{wen:2017}. 
    \item \fullmodel Our full model, where we first pre-train sentence representations $z^*_t$ to maximize the likelihood $ p(x_{t+1}, a_t |\mathcal{A}, z^*_t, z_{0:t-1})p_z(z^*_t | \mathcal{A}, x_t)$, and then we train models to predict $p_{x}(x_t | z^*_t, x_{0:t-1})$ and $\hat{p}_{z}(z_t |\mathcal{A}, v, x_{0:t-1})$.
\end{itemize}
To focus the evaluation on the linguistic and strategic aspects of the dialogue, all systems use the same model for predicting the final agreement represented by the dialogue, which is implemented as a bidirectional GRU with attention over the words of the dialogue.


\subsection{Likelihood Models}
First, we experiment with models using no RL or rollouts.

\subsubsection{Perplexity}
Models were developed to maximize the likelihood of human dialogues, which is an indicator of how human-like the language is (we observed qualitatively that the two were strongly correlated). Results are shown in Table \ref{table:perplexity}.

The use of a hierarchical RNN model improves performance over a strong baseline from previous work.

Perhaps surprisingly, our hierarchical latent-variable model is also able to achieve state-of-the-art performance. 
This shows our model's discrete encodings of messages are as informative for predicting the future dialogue as the more-expressive embeddings used by the hierarchical baseline.

\begin{table}
\centering
\label{my-label}
\begin{tabular}{@{}lccc@{}}
\toprule
Model & \begin{tabular}[c]{@{}l@{}}Validation\\ Perplexity\end{tabular} & \begin{tabular}[c]{@{}l@{}}Test\\ Perplexity\end{tabular} &  \\ 
\midrule
\rnn & 5.62 & 5.47 \\
\hierarchical & 5.37 & 5.21\\
\latentintents \hspace{8mm} & 5.61 & 5.46 \\
\fullmodel & 5.37 & 5.24\\
\bottomrule
\end{tabular}
\caption{\label{table:perplexity} Likelihood of human dialogues using different models. Our model with discrete message representations is able to achieve state-of-the-art performance, showing that the representations capture relevant aspects of messages for predicting the future dialogue. The size of 95\% CI is within 0.03 for each entry. }

\end{table}






\begin{table}[]
\centering

\begin{tabular}{@{}lccc@{}}
\toprule
Model & \begin{tabular}[c]{@{}l@{}}Score vs. \\ \rnn \end{tabular} & \begin{tabular}[c]{@{}l@{}}Score vs. \\ \hierarchical\end{tabular} & \\
\midrule
\rnn & 5.33 & 5.17 \\
\hierarchical& 5.37 & 5.08 \\
\latentintents \hspace{1mm} & 4.68 & 4.66 \\
\fullmodel& 6.75 & 6.57 \\

\bottomrule
\end{tabular}
\caption{\label{table:supervised:reward}Comparison of different models based on their end-task reward. Our clusters substantially improve reward, indicating that they make it easier for supervised learning to model strategic decision making. The size of 95\% CI is within 0.14 for each entry.}
\end{table}

\subsubsection{Coherence of Clusters}
Table \ref{table:clusters} shows random samples of messages generated by different clusters from our predicted state model, and the \latentintents model.

Qualitatively, the states from our model show a higher degree of semantic coherence, and higher linguistic variability. Compared to the \latentintents, our approach tends to generate more dissimilar surface strings, but with more similar semantics. 
Our clusters appear to capture \emph{meaning} rather than \emph{form}.

\begin{table}[t]
\centering
\begin{tabular}{@{}lccc@{}}
\toprule
\scalebox{0.90}{Rollout Type} & \scalebox{0.90}{\begin{tabular}[c]{@{}l@{}}Score vs. \\ \textsc{No Rollouts}\end{tabular}} & \scalebox{0.90}{\begin{tabular}[c]{@{}l@{}}Score vs. \\ \textsc{Baseline Rollouts}\end{tabular}}  \\
\midrule
\textsc{No Rollouts}  & 5.08  & 4.91   \\
\textsc{Baseline} &  7.81 & 6.57  \\
\textsc{Diverse} & 8.41  & 7.36   \\
\bottomrule
\end{tabular}
\caption{\label{table:rollouts} Comparison of different rollout strategies for the \fullmodel. \textsc{Diverse} rollouts use distinct latent variables to create more semantic diversity in rollout candidates, significantly improving performance. The size of 95\% CI is within 0.19 for each entry.}

\end{table}

\begin{table*}[t]
\centering
\begin{tabular}{cll}
\toprule
\scalebox{0.84}{Cluster}& \scalebox{0.84}{\latentintents} & \scalebox{0.84}{\fullmodel}\\
\midrule
\scalebox{0.84}{1}&  \scalebox{0.84}{i can give you the books but , i would need the hat and the balls} & \scalebox{0.84}{i would like the hat and 1 book} \\
&  \scalebox{0.84}{i can do that . i need both balls and one book} & \scalebox{0.84}{i can't give up the hat , but i can offer you the book and 2 balls} \\
\hline
\scalebox{0.84}{2}& \scalebox{0.84}{i need both books and the hat} & \scalebox{0.84}{i want the hat} \\
& \scalebox{0.84}{how about you get the hat and 1 ball} & \scalebox{0.84}{i need the hat . you can have all the books and the balls} \\
\hline
\scalebox{0.84}{3}& \scalebox{0.84}{i can not make that deal . i need the hat and one book} & \scalebox{0.84}{i can give you the hat and 1 ball}\\
& \scalebox{0.84}{i can give you the hat and 1 ball} & \scalebox{0.84}{i would like the books and a ball}\\
\hline
\scalebox{0.84}{4}& \scalebox{0.84}{i need two books and the hat} & \scalebox{0.84}{i need the books and the hat}\\
& \scalebox{0.84}{i need the hat , you can have the rest} & \scalebox{0.84}{i can give you the balls but i need the hat and books}\\
\hline
\scalebox{0.84}{5}& \scalebox{0.84}{i can give you the hat if i can have the rest} & \scalebox{0.84}{could i have the books and a ball ?}\\
& \scalebox{0.84}{i want one of each} & \scalebox{0.80}{i would like the books and one ball}\\
\bottomrule                          
\end{tabular}

\caption{\label{table:clusters} Messages sampled from different clusters, where \textbf{2 books}, \textbf{1 hat}, and \textbf{2 balls} are available. Our method's clusters are much more semantically coherent than the baseline, and correspond to different ways of proposing the same deal.
} 
\end{table*}

\subsubsection{End Task Performance}
We measure the performance of the different models on their end-task reward over 1000 negotiations in self-play. Results are shown in Table \ref{table:supervised:reward}.
We find that the use of our latent representations leads to a large improvement in the reward, indicating that our representations make it easier for the supervised model to learn the latent decision making process in the human dialogues it was trained on.

\begin{table}[]
\centering
\begin{tabular}{@{}lcccc@{}}
\toprule
Model & \begin{tabular}[c]{@{}l@{}}Score vs. \\ \human \end{tabular} & \begin{tabular}[c]{@{}l@{}}Language \\ quality \end{tabular} & \begin{tabular}[c]{@{}l@{}}Number \\ of turns \end{tabular} \\
\midrule
\textsc{Full + Rollout} \hspace{2.5mm} & 7.45 & 3.55 & 4.89\\
\textsc{RNN + Rollout} & 6.99 & 3.43 & 4.38 \\
\midrule
\textsc{Full + RL} & 6.26 & 3.60& 6.52 \\
\textsc{RNN + RL} & 6.01 & 3.52 & 3.99 \\
\midrule
\fullmodel & 5.42 & 3.68 & 3.07 \\
\rnn & 5.30 & 3.56 & 3.96 \\
\midrule
\human & 6.64 & 3.85 &  6.36 \\
\bottomrule
\end{tabular}
\caption{\label{table:human_eval} Performance of our \fullmodel model and the highly optimized \rnn model against humans. In all cases, our \fullmodel model achieves both higher scores and uses higher quality language than \rnn.  }
\end{table}
\subsection{Hierarchical Planning}
Next, we evaluate different rollout strategies:

\begin{itemize}
\item \textsc{Baseline Rollouts} following \citet{lewis:2017}, where first $K$ candidate sentences are sampled from the model, and then tokens are sampled iteratively from $p_x$ until reaching the end of the dialogue. 
\item \textsc{Diverse Rollouts} where we first choose the top $K$ unique $z_t$ from $\hat{p}_z$. By choosing unique $z_t$ we aim to increase the semantic diversity of the candidates.
\end{itemize}
We evaluate compared to the baseline model and word-level rollouts and record the average score.
Results are shown in Table \ref{table:rollouts}, and that the \textsc{Diverse Rollouts} that use our message representations lead to a large improvement over previous approaches.

\subsection{Finetuning with Reinforcement Learning}
A challenge in using RL for end-to-end text generation models is that optimising for reward can adversely affect language generation. In selfplay, the model can learn to achieve a high reward by finding uninterpretable sequences of tokens that the baseline model was not exposed to at training time. We compare several RL approaches:
\begin{itemize}
    \item \textsc{All-RL} Reinforcement learning after pre-training with supervised learning.
    \item \textsc{All-RL+SV} Interleaved RL and supervised learning updates, weighting supervised updates with a hyperparameter $\alpha$, similarly to \citet{lewis:2017}.
    \item \textsc{Pred-RL} Reinforcement learning only to fine-tune the intent model $\hat{p}_{z}$, with all other parameters fixed.
\end{itemize}

We measure both the average reward of the model (a measure of its ability to achieve its goals) and the perplexity of the model on human dialogues (a measure of how human-like the language is). After hyper-parameter search, we plot the reward of the best model whose perplexity is at most $a$.

\begin{figure}[t]
\includegraphics[width=\columnwidth]{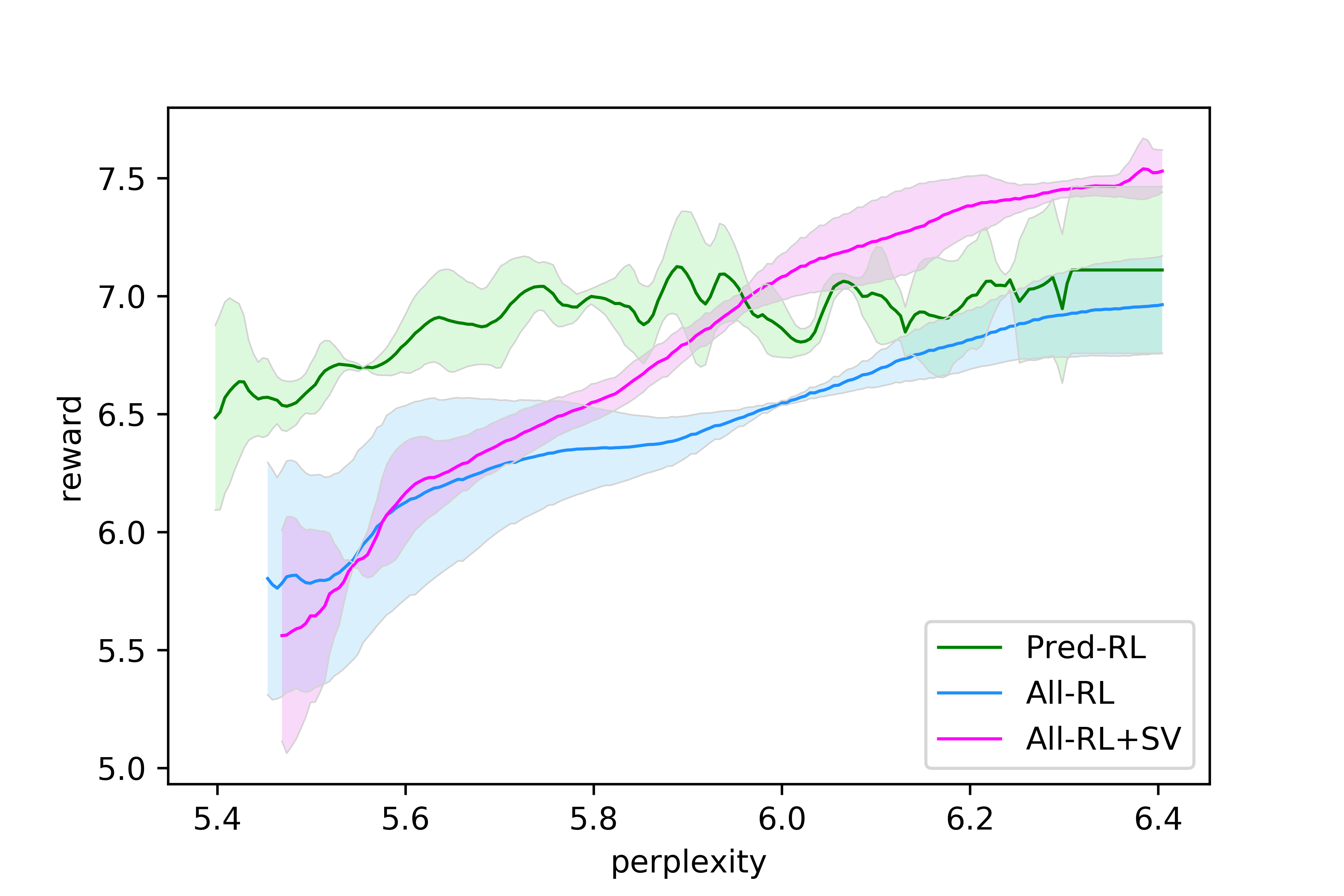}
\caption{\label{figure:score_vs_perplexity} Plotting reward against language quality (lower perplexity is better) during reinforcement learning training, in dialogues with the \hierarchical model. Our method (green) achieves higher rewards while maintaining human-like language (top left of graph).} 
\end{figure}

Results are shown in Figure \ref{figure:score_vs_perplexity}.
Using RL on all parameters allows high rewards at the price of poor quality language.
Only fine-tuning $\hat{p}_z$ allows the model to improve its strategic decision making, while retaining human-like language.

\subsection{Human Evaluation}
To confirm our empirical results, we evaluate our model in dialogues with people. We ran 1415 dialogues on MTurk, where humans were randomly paired with either one of the models or another human. We then asked humans to rate the language quality of their partner (from 1 to 5). Results are shown in Table \ref{table:human_eval}. We observe that our model consistently outperforms the baseline model \citep{lewis:2017} both in the end-task reward and the language quality.

\section{Analysis}
\label{section:analysis}
Results in section \ref{section:experiments} show quantitatively that our hierarchical model improves the likelihood of human generated language and the average score achieved by the agent.
Here, we investigate specific issues that the model improved on, and identify remaining challenges.
We analyzed 1000 dialogues between our \fullmodel and the \hierarchical baseline. These models achieve similar perplexity on human dialogues (Table \ref{table:perplexity}).


\subsection{Linguistic Diversity}
First, we measure the diversity of the agents' language.

RNN language models are known to prefer overly generic messages. In our task, this often manifests itself as short messages such as \emph{deal} or \emph{ok}. We measure the frequency of simple variations on these messages, and find that the \hierarchical model uses generic messages far more often than \fullmodel (815 times vs. 245).

The messages sent by \fullmodel are also longer on average (8.9 words vs. 6.7, ignoring the end-of-dialogue token), giving further evidence of greater complexity.

We also find that the \fullmodel is substantially more creative in generating new messages beyond those seen in its training data. In total, \fullmodel sends 875 unique message strings, of which 525 (60\%) do not appear in the training data. In contrast, \hierarchical sends fewer unique message strings (751), and just 18\% of these are not copied from the training data.

\subsection{Self-consistency of Messages}
Models can output inconsistent messages, such as \emph{I really need the hat. I can give you the hat and one ball}. We searched for messages that mentioned the same item type multiple times, and then manually evaluated whether it was consistent. The \fullmodel model was more prone to this error than \hierarchical (23 times vs. 11), though this fact may be a consequence of its greater creativity, and the problem only occurred in roughly 1\% of messages.

\subsection{Consistency with Input}
We also investigate whether messages are consistent with the context---for example, models may emit messages such as \emph{I'd like the hat and books; you keep the 3 balls} when there are not 3 balls available. We use simple pattern matching for several such errors, and found that the \fullmodel performed slightly better (15 errors vs. 19).

\subsection{Consistency with Dialogue Context}
\citet{lewis:2017} describe cases where an agent indicates it is simply re-stating an agreement, when it is actually proposing a new deal (e.g. \emph{you get 2 hats / Okay deal, so I get 3 hats}). 
Interestingly, we found this behaviour only happened with the models using rollouts. 
While this tactic is effective against our models, it would be frustrating for humans, and future work should address this issue.

\subsection{Repetitiveness}
Previous work noted that reinforcement learning models were prone to an extortion tactic of simply repeating the same demand until acceptance. We measured how often agents repeated the same message in a dialogue, comparing the \textsc{All-RL+SV}  model based on previous work, with our \textsc{Pred-RL} model. Our model was substantially less repetitive: only 1\% of dialogues contained a repetition of the same message, compared to 12\% for the baseline.


\newcommand{\itemvalue}[3]{#2x\textbf{#1} \textit{value}=#3}
\newcommand{\deal[3]}

\begin{figure}
\centering
\scalebox{0.70}{%
\begin{tabular}{|llc|}
\hline
\textbf{Input} &&\\
\textsc{Diverse Rollouts} & \multicolumn{2}{l|}{\itemvalue{book}{1}{9}  \itemvalue{hat}{1}{1}  \itemvalue{ball}{4}{0}}  \\
\textsc{Baseline Rollouts}& \multicolumn{2}{l|}{\itemvalue{book}{1}{0}  \itemvalue{hat}{1}{6}  \itemvalue{ball}{4}{1} }\\
\hline
\textsc{Diverse Rollouts}    &  \multicolumn{2}{l|}{I will take the book and hat and you can have the balls.} \\
\textsc{Baseline Rollouts} &  \multicolumn{2}{l|}{I need the hat and two balls}  \\ 
\textsc{Diverse Rollouts}    &  \multicolumn{2}{l|}{The balls are worthless, I need the hat and the book.} \\
\textsc{Baseline Rollouts} & \multicolumn{2}{l|}{ I need the hat or no deal} \\ 
\textsc{Diverse Rollouts}    &  \multicolumn{2}{l|}{Then no deal.} \\
\textsc{Baseline Rollouts} & \multicolumn{2}{l|}{What about the balls?} \\
\textsc{Diverse Rollouts}    &  \multicolumn{2}{l|}{You can have the hat but I need the book.} \\
\textsc{Baseline Rollouts} & \multicolumn{2}{l|}{ How about I get the balls and 1 hat?} \\ 
\textsc{Diverse Rollouts}    &  \multicolumn{2}{l|}{Ok} \\
\hline
\textbf{Output} & & \textbf{Reward}\\
\textsc{Diverse Rollouts} & 1x\textbf{book} & 9/10\\
\textsc{Baseline Rollouts} &  1x\textbf{hat} 4x\textbf{ball} & 10/10 \\
\hline
\end{tabular}
}

\caption{\label{figure:stubborn}Dialogue between two models using different types of rollouts. The \textsc{Diverse Rollouts} model makes several attempts to win the \emph{hat}, before compromizing.
}
\end{figure}
\section{Related Work}

Traditional goal-orientated dialogue models have first generated symbolic \emph{intents}, capturing the meaning of the message, and then generated text to match the intent (e.g. \citet{williams:2007}, \citet{keizer:2017}). Our approach can be seen as a latent model for generating intents.
Our model is most closely related to other recent latent variable hierarchical dialogue models from \citet{serban:2016c}, \citet{wen:2017} and \citet{cao:2017}. An important difference is that both these approaches optimize latent representations $z$ to maximize the likelihood of generating the next message---whereas our model pretrain's $z$ to maximize the likelihood of the continuation of the dialogue, to better capture the semantics of the message rather than its surface form. While other ways of learning discrete latent representations were proposed recently \citep{oord:2017, kaiser:2018}, we have shown that our approach leads to higher performance on a strategic dialogue task.

Other work has explored generating sentence embeddings for open domain text---for example, based on maximizing the likelihood of surrounding sentences \citep{kiros:2015}, supervised entailment data \citep{conneau:2017}, and auto-encoders \citep{bowman:2015}.

\section{Conclusion}
We have introduced a novel approach to creating sentence representations, within the context of an end-to-end strategic dialogue system, and have shown that our hierarchical approach improves text generation and planning.
We identified a number of challenges faced by previous work, and show empirically that our model improves on these aspects.
Future work should apply our model to other dialogue settings, such as cooperative strategic dialogue games \cite{he:2017}, or multi-sentence generation tasks, such as long document language modelling \cite{merity:2016}.

\bibliography{references}
\bibliographystyle{icml2018}

\end{document}